\PassOptionsToPackage{unicode}{hyperref}
\PassOptionsToPackage{hyphens}{url}
\PassOptionsToPackage{dvipsnames,svgnames,x11names}{xcolor}
\documentclass[
  11pt,
]{article}
\usepackage{xcolor}
\usepackage[paper=a4paper,margin=1in]{geometry}
\usepackage{amsmath,amssymb}
\usepackage{iftex}
\ifPDFTeX
  \usepackage[T1]{fontenc}
  \usepackage[utf8]{inputenc}
  \usepackage{textcomp} 
\else 
  \usepackage{unicode-math} 
  \defaultfontfeatures{Scale=MatchLowercase}
  \defaultfontfeatures[\rmfamily]{Ligatures=TeX,Scale=1}
\fi
\usepackage{lmodern}
\ifPDFTeX\else
\fi
\IfFileExists{upquote.sty}{\usepackage{upquote}}{}
\IfFileExists{microtype.sty}{
  \usepackage[]{microtype}
  \UseMicrotypeSet[protrusion]{basicmath} 
}{}
\makeatletter
\@ifundefined{KOMAClassName}{
  \IfFileExists{parskip.sty}{%
    \usepackage{parskip}
  }{
    \setlength{\parindent}{0pt}
    \setlength{\parskip}{6pt plus 2pt minus 1pt}}
}{
  \KOMAoptions{parskip=half}}
\makeatother
\usepackage{longtable,booktabs,array}
\usepackage{calc} 
\usepackage{etoolbox}
\makeatletter
\patchcmd\longtable{\par}{\if@noskipsec\mbox{}\fi\par}{}{}
\makeatother
\IfFileExists{footnotehyper.sty}{\usepackage{footnotehyper}}{\usepackage{footnote}}
\makesavenoteenv{longtable}
\usepackage{graphicx}
\makeatletter
\newsavebox\pandoc@box
\newcommand*\pandocbounded[1]{
  \sbox\pandoc@box{#1}%
  \Gscale@div\@tempa{\textheight}{\dimexpr\ht\pandoc@box+\dp\pandoc@box\relax}%
  \Gscale@div\@tempb{\linewidth}{\wd\pandoc@box}%
  \ifdim\@tempb\p@<\@tempa\p@\let\@tempa\@tempb\fi
  \ifdim\@tempa\p@<\p@\scalebox{\@tempa}{\usebox\pandoc@box}%
  \else\usebox{\pandoc@box}%
  \fi%
}
\def\fps@figure{htbp}
\makeatother
\providecommand{\tightlist}{%
  \setlength{\itemsep}{0pt}\setlength{\parskip}{0pt}}
\usepackage{kotex}
\usepackage{microtype}
\usepackage{setspace}
\usepackage{titlesec}
\usepackage{caption}
\usepackage{float}
\usepackage{titling}
\usepackage{xurl}
\usepackage{fvextra}
\usepackage{needspace}

\ifPDFTeX
  \DeclareUnicodeCharacter{2192}{\ensuremath{\rightarrow}}
  \DeclareUnicodeCharacter{2212}{\ensuremath{-}}
\fi

\definecolor{PaperBlue}{HTML}{1F4E79}

\pretitle{\begin{center}\Large\bfseries}
\posttitle{\par\end{center}\vspace{-0.6em}}
\preauthor{\begin{center}\normalsize}
\postauthor{\par\end{center}\vspace{-0.6em}}
\predate{\begin{center}\small}
\postdate{\par\end{center}\vspace{0.8em}}

\titleformat{\section}
  {\large\bfseries}
  {}{0pt}{}
\titleformat{\subsection}
  {\normalsize\bfseries}
  {}{0pt}{}
\titleformat{\subsubsection}
  {\normalsize\itshape}
  {}{0pt}{}

\titlespacing*{\section}{0pt}{1.25em}{0.55em}
\titlespacing*{\subsection}{0pt}{0.9em}{0.35em}
\titlespacing*{\subsubsection}{0pt}{0.75em}{0.25em}

\newlength{\abstractrulewidth}
\AtBeginEnvironment{abstract}{%
  \small
  \begin{center}
  \rule{\abstractrulewidth}{0.4pt}
  \end{center}
}

\AtEndEnvironment{abstract}{%
  \vspace{-1.8em}
  \begin{center}
  \rule{\abstractrulewidth}{0.4pt}
  \end{center}
  \vspace{-0.3em}
}

\setkeys{Gin}{width=\linewidth,height=0.48\textheight,keepaspectratio}

\AtBeginEnvironment{tabular}{\small}
\AtBeginEnvironment{longtable}{\small}
\DefineVerbatimEnvironment{Highlighting}{Verbatim}{breaklines,breakanywhere,commandchars=\\\{\}}
\RecustomVerbatimCommand{\VerbatimInput}{VerbatimInput}{breaklines,breakanywhere}
\usepackage{bookmark}
\IfFileExists{xurl.sty}{\usepackage{xurl}}{} 
\makeatletter
\@ifundefined{xmpquote}{}{}
\makeatother
\hypersetup{
  pdftitle={AI\_LectureNote: A Retrospective Pilot Study of a Post-ASR Workflow for English-Script Rendering and Semantic Drift in Korean--English Medical Lectures},
  colorlinks=true,
  linkcolor={PaperBlue},
  filecolor={Maroon},
  citecolor={PaperBlue},
  urlcolor={PaperBlue},
  pdfcreator={LaTeX via pandoc}}

\title{AI\_LectureNote: A Retrospective Pilot Study of a Post-ASR
Workflow for English-Script Rendering and Semantic Drift in
Korean--English Medical Lectures}
\author{
\vspace{1.3em}
Kyeongeon Lee, Donghoon Chang, Seungryeol Baek,\\
Taehong Kim, Wonjun Yang\\[0.7em]
\small Sungkyunkwan University, Republic of Korea\\[0.5em]
\scriptsize
\texttt{boy.skier@g.skku.edu} \quad
\texttt{dhchang5765@gmail.com} \quad
\texttt{bsr20030507@g.skku.edu}\\
\texttt{jinanara@skku.edu} \quad
\texttt{judereal15@gmail.com}\\[0.2em]
\scriptsize 
}
\date{}

\begin{document}
\maketitle
\vspace{-2.5em}
\begin{abstract}
AI\_LectureNote is a historical, readability-oriented post-ASR workflow
for Korean--English medical lectures. It rewrites speech-to-text output
into study transcripts while restoring Latin-script medical terms rather
than Korean phonetic transliterations. We retrospectively evaluate the
workflow on four author-recorded lectures across five conditions. In
this pilot, post-processing raised the macro English-script rendering
rate from 0.39 to 0.71 on the whisper-1 path and from 0.26 to 0.65 when
applied to 3-minute chunked gpt-4o-transcribe output. However,
English-script rendering did not imply semantic faithfulness: the two
post-processed conditions showed semantic drift in 34 and 36 of 282
reference sentences and polarity failures in 11 and 13 of 101
polarity-cue rows. A descriptive cross-input comparison suggested
different candidate failure patterns: polarity-failure sets overlapped
more strongly across front-ends (Jaccard 0.60; 9 shared of 15 unioned
failures) than general semantic-drift sets (Jaccard 0.23; 13 shared of 57
unioned drifts). This
single-annotator pilot documents concrete failure modes rather than
population rates and supports evaluating surface accuracy, term-script
rendering, chunk-level script consistency, and medical-meaning
preservation separately.
\end{abstract}

\section{1. Introduction}\label{introduction}

Students using recorded Korean--English medical lectures often need
something different from a verbatim acoustic transcript. A useful study
transcript should keep the lecturer's medical claims readable, render
English medical terminology in English script, and retain clinically or
biologically important directionality. Generic ASR metrics do not
directly measure this target. In Korean--English medical lectures, the
gap is especially visible when ASR renders embedded English terms as
Korean phonetic strings (for example, ``carbonic anhydrase inhibitor''
may be rendered as ``카르보닉 아나이드레이즈 이니비터''),  producing fluent-looking but
hard-to-study transcripts.

AI\_LectureNote was built for that practical problem. It was a
student-built prototype that used the OpenAI \texttt{whisper-1} API as
its raw ASR front-end and added domain-aware post-processing to restore
medical terms and regularize wording into a study transcript. The system
originated as an internal study aid. Here, we re-run that historical
post-processing code unchanged and evaluate it retrospectively.

The central distinction is between readability and faithfulness.
English-script rendering makes a transcript easier to scan, search, and
align with textbooks, but it can hide different errors. A rewritten
transcript may replace a Korean phonetic string with an English medical
word that looks plausible while being the wrong word. It may compress an
explanation and drop the causal relation, or it may flip a polarity cue
such as hypo- versus hyper-. These are not only formatting errors; they
alter what a student would study.

The motivating question is therefore not whether AI\_LectureNote is a
better ASR system than a modern commercial transcriber. The question is
narrower: when readability-oriented post-ASR rewriting improves
English-script rendering, what happens to medical-meaning faithfulness?
This question remains relevant even with newer raw STT front-ends such
as gpt-4o-transcribe, because fluent raw output and readable
post-processed output can fail in different ways.

We compare five conditions over the same four author-recorded lectures.
Two conditions are the historical path and its raw front-end: raw
whisper-1 and AI\_LectureNote. Two are gpt-4o-transcribe conditions: raw
chunked output and the same chunking with a single minimal Korean prompt
asking for English medical terms in English. The fifth condition,
\texttt{gpt4o\_ailn\_post}, applies the same AI\_LectureNote
post-processing to raw gpt-4o-transcribe output. This cross-input
condition changes the upstream raw transcript while holding the
downstream post-processing stage fixed.

The pilot reveals a readability--faithfulness trade-off: post-processing
improves English-script rendering, while observed semantic and polarity
errors require separate evaluation of medical meaning.

Accordingly, this report treats AI\_LectureNote as a diagnostic case
study in the trade-off between readability-oriented post-ASR rewriting
and medical-meaning faithfulness, rather than as a benchmark of ASR
systems. Detailed non-claims are handled in Limitations; the main scope
sentence is simple: this is a four-lecture, five-condition, single
author-annotator pilot of one historical workflow.

Our contributions are:

\begin{enumerate}
\def\labelenumi{\arabic{enumi}.}
\tightlist
\item
  A five-condition retrospective protocol for evaluating a
  Korean--English medical study-transcript workflow, including a
  cross-input control that applies the same post-processing to a
  different raw ASR front-end.
\item
  Two readability-oriented metrics --- English-script rendering and
  chunk-level script consistency --- reported separately from medical
  correctness.
\item
  An operational labeling scheme for post-processed errors as introduced
  when raw ASR preserved the relevant information but post-processing
  lost or changed it, and propagated when post-processing carried
  forward an upstream ASR error.
\end{enumerate}

\section{2. Related Work}\label{related-work}

\textbf{Code-switched and medical ASR.} Code-switching between a matrix
language and embedded English is a known challenge for ASR, especially
when the system must choose between native-script transliteration and
English-script rendering (Sitaram et al., 2019; Wang et al., 2019; Paik
et al., 2025). Medical lecture speech adds technical terms,
abbreviations, and morphology that are sparse in general-purpose
training data. Recent medical ASR work similarly reports that domain
terminology remains difficult even when general transcription fluency is
high (Adedeji et al., 2024, 2025). Korean--English medical lectures
combine these problems: the sentence frame may be Korean, while many
target concepts are conventionally learned and searched in English.

\textbf{LLM post-ASR correction.} A common response to raw ASR errors is
to place a language model after the transcriber to clean, normalize, or
correct the transcript. Recent work has explored LLM-based correction
pipelines for full-text speech recognition and specialized speech
settings (Ma et al., 2023; Tang et al., 2025; Zheng et al., 2026). Such systems can
improve readability and reduce surface noise, but an unconstrained
rewriting stage may also paraphrase, compress, substitute, or
hallucinate content. This risk is easy to miss if evaluation stops at
WER or at a term-recall score. AI\_LectureNote is a retrospective
example of this pattern: a readability-oriented LLM rewriting stage
whose gains and costs can be measured separately.

\textbf{Domain-term rendering and normalization.} Technical
transcription often benefits from contextual biasing, hotword lists, or
dynamic vocabulary methods that encourage systems to retain or
canonicalize domain terms (Sudo et al., 2024a, 2024b). These methods
motivate a more constrained alternative to free-form rewriting. They
also clarify why this paper uses ``English-script rendering'' rather
than ``preservation'' for the headline readability metric. A curated
reference term is counted when it appears in English/Latin script in the
output. This is intentionally narrower than term correctness: a wrong
English term can still be rendered in English script, and a hallucinated
English term may not be penalized directly by recall.

\textbf{Safety-critical polarity and negation errors.} Medical text is
sensitive to directionality and assertion status. Negation, uncertainty,
and polarity cues have long been treated as first-class targets in
clinical NLP because presence versus absence, increase versus decrease,
and hypo- versus hyper- can change the meaning of a medical statement
(Chapman et al., 2001; Uzuner et al., 2011; Peng et al., 2017). Lecture
transcripts are not clinical records, but the same linguistic issue
applies to study materials. A transcript that flips hypokalemia to
hyperkalemia or omits a directional cue is not merely less polished; it
teaches a different claim. For this reason, polarity is annotated
separately from general semantic drift.

\section{3. System Retrospective:
AI\_LectureNote}\label{system-retrospective-ai_lecturenote}

AI\_LectureNote is a post-ASR, study-transcript generation workflow with
the pipeline:

\begin{verbatim}
audio→raw ASR (whisper-1)→domain-aware LLM post-processing→study transcript
\end{verbatim}

After raw STT, AI\_LectureNote  operates on text:
it chunks the raw transcript, applies a correction pass for abnormal or disfluent
wording, and then applies an ``Englishing'' pass that converts Korean-phonetic
renderings of medical English terms into English/Latin script while keeping
Korean sentences in Korean. The target output is therefore a
readability-oriented study transcript rather than a verbatim conversational
transcript. The workflow could optionally build a knowledge graph from the
resulting study transcript, but this stage was not enabled for the present
evaluation.

This design motivates the faithfulness analysis. Because the post-processing
stage is generative rewriting rather than constrained glossary replacement, it
can improve English-script rendering while introducing substitutions, omissions,
compression, or polarity changes. We therefore evaluate AI\_LectureNote  as a
readability-versus-faithfulness trade-off rather than as raw transcription
accuracy.

Two historical caveats matter for interpretation. This study does not
evaluate the original production corpus; instead, it re-runs the
historical code with unchanged prompts and configuration on four
author-recorded lectures. The whisper-1 path was re-executed in 2026, so
it should not be read as a 2023 snapshot. Because whisper-1 may have
changed internally, we report model names and execution dates but do not
claim byte-for-byte historical reproduction.

\section{4. Pilot Materials and Evaluation
Setup}\label{pilot-materials-and-evaluation-setup}

\subsection{4.1 Lectures and conditions}\label{lectures-and-conditions}

The pilot uses four Korean--English medical lectures recorded specifically for this study by the authors, 
all delivered extemporaneously from loose AI-generated topic outlines without pre-written scripts or source texts.
The recordings contain no patient data and are illustrative teaching content whose medical accuracy is not guaranteed; all speaker voices are from the author-speakers, who consented to research release.
\Needspace{10\baselineskip}
\textbf{Table 1. Lecture recordings used in the pilot.}

\begin{center}
\small
\begin{tabular}{@{}llrl@{}}
\toprule\noalign{}
lecture & topic & duration & speaker \\
\midrule\noalign{}
\texttt{diuretics\_01} & Diuretics & 13m 13s & lke \\
\texttt{acuteinflammation\_02} & Acute inflammation & 13m 7s & lke \\
\texttt{anthrax\_01} & Anthrax (\emph{Bacillus anthracis}) & 12m 16s &
cdh \\
\texttt{anticancerdrugs\_02} & Anticancer drugs & 4m 30s & cdh \\
\bottomrule\noalign{}
\end{tabular}
\end{center}

Three lectures formed the initial pilot set;
\texttt{anticancerdrugs\_02} was added later as a shorter same-speaker
extension.

From the same audio we produced five output conditions:

\begin{enumerate}
\def\labelenumi{\arabic{enumi}.}
\tightlist
\item
  \textbf{raw whisper-1} --- the historical pipeline's raw ASR stage
  (whisper-1; Radford et al., 2023), re-executed in 2026 with
  per-lecture API dates recorded in the manifest.
\item
  \textbf{AI\_LectureNote} --- whisper-1 output passed through the
  historical AI\_LectureNote post-processing (code unchanged).
\item
  \textbf{raw gpt-4o-transcribe} --- the OpenAI gpt-4o-transcribe STT
  front-end (gpt-4o-transcribe; OpenAI, 2025) as accessed for this
  study, transcribed in 3-minute non-overlapping chunks.
\item
  \textbf{prompted gpt-4o-transcribe} --- the same chunking with a
  single Korean prompt (``의학 강의 녹음입니다. 영어 의학 용어는 영어로
  표기합니다.'' --- \emph{``This is a medical lecture recording. Write
  English medical terms in English.''}), included only as a preliminary
  prompt-sensitivity check.
\item
  \textbf{gpt4o\_ailn\_post} --- raw gpt-4o-transcribe output from
  condition 3 passed through the same AI\_LectureNote post-processing
  stage as condition 2. This condition serves as a cross-input
  diagnostic control rather than an optimized modern pipeline.
\end{enumerate}

For space, some figures abbreviate these condition names: raw
gpt-4o-transcribe appears as ``gpt-4o raw'', and the two post-processed
conditions (AI\_LectureNote and \texttt{gpt4o\_ailn\_post}) appear as
``whisper→AILN'' and ``gpt4o→AILN post''.

\begin{figure}
\centering
\pandocbounded{\includegraphics[keepaspectratio,alt={Evaluation design. Four author-recorded lectures were processed into five outputs and scored against author-created study-transcript references. The two post-processed conditions use the same historical AI\_LectureNote rewriting stage.}]{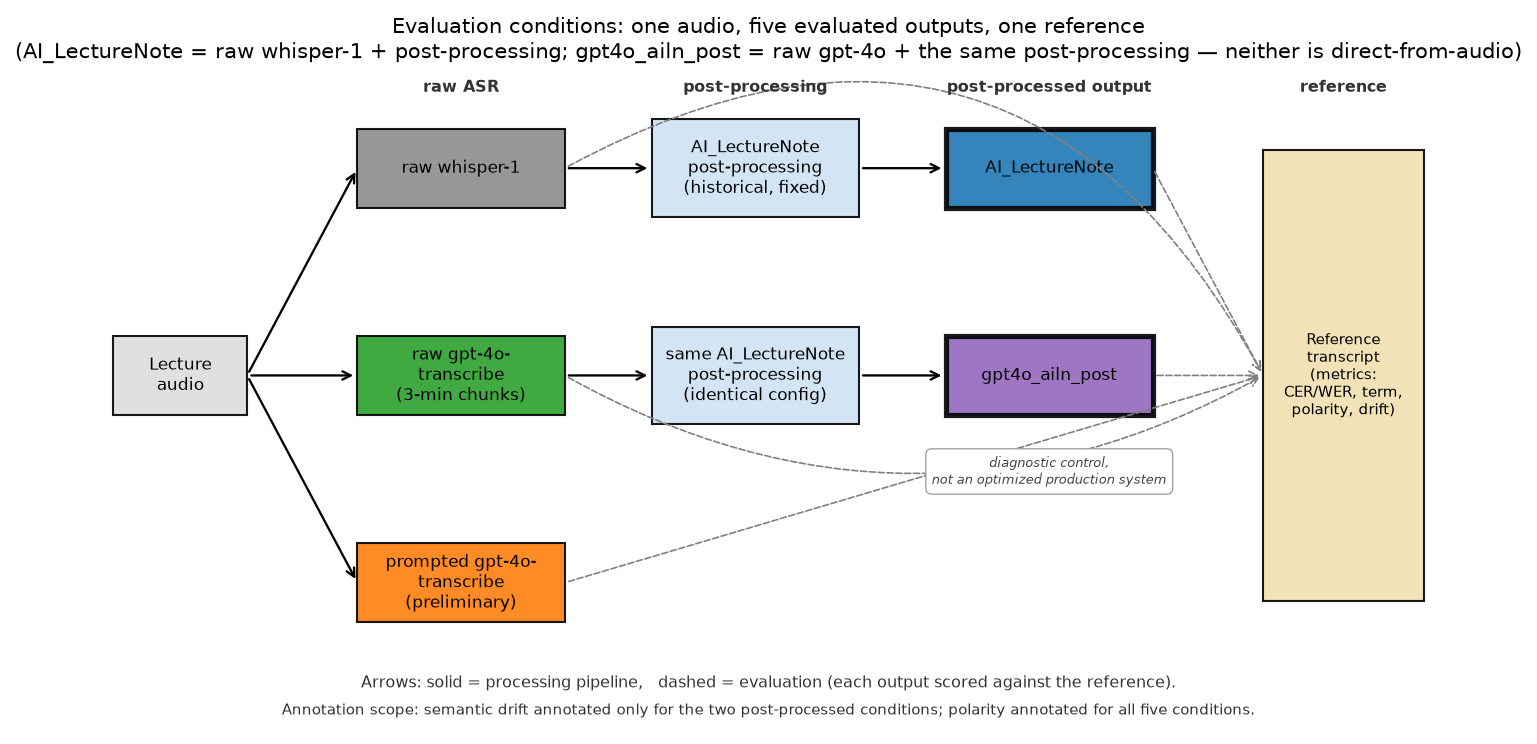}}
\caption{Evaluation design. Four author-recorded lectures were processed
into five outputs and scored against author-created study-transcript
references. The two post-processed conditions use the same historical
AI\_LectureNote rewriting stage.}
\end{figure}

\subsection{4.2 Reference policy}\label{reference-policy}

The references are author-created, lightly cleaned, content-preserving
study transcripts, not strict verbatim conversational transcripts or
independent gold-standard ASR references. They preserve medical content,
English medical terms, polarity/direction cues, and medically imperfect
speaker statements, while lightly regularizing non-content artifacts
such as fillers, coughs, abandoned fragments, and immediate stutter
repetitions when this did not change the medical claim. Therefore,
CER/WER should be interpreted relative to a study-transcript target
rather than pure conversational-ASR accuracy. Rendering and
semantic/polarity analyses are less sensitive to fillers and spacing but
still depend on this reference policy. Term canonicalization is handled
in annotation, not by editing the reference. The recordings and
references are research materials for evaluating transcription and
post-ASR processing; they are not medical or educational guidance, and
the medical accuracy of the spoken lecture content was not independently
verified.

\section{5. Metrics and Annotation}\label{metrics-and-annotation}

\subsection{5.1 Metric definitions}\label{metric-definitions}

Normalization and metric definitions were frozen before analysis:
English text is lowercased, punctuation is mapped to spaces, and
whitespace is collapsed; polarity morphemes (hypo/hyper,
acidosis/alkalosis) are never merged. The primary surface metric is
\textbf{CER (character error rate)} computed on whitespace-stripped
characters (removing the effect of Korean spacing). \textbf{WER (word
error rate)} is reported as a secondary metric on normalized tokens, but
is less interpretable for paraphrased study transcripts.

The headline readability metric is the \textbf{English-script rendering
rate}: capped recall of curated reference domain-term occurrences
rendered in English/Latin script. It measures script occurrence, not
whether the rendered term is medically correct in the correct sentence,
and it does not directly penalize hallucinated English-script terms
except through missing reference terms. The \textbf{Korean phonetic
count} is the pooled count of curated reference domain terms rendered in
Korean phonetic/transliterated script, summed over lectures. Per-lecture
domain-term lists were initialized by automatically extracting
Latin-script term candidates from the reference transcripts. The author
curated the kept term set, canonical English forms, accepted English
variants, and polarity tags. Korean phonetic variants were
machine-assisted from observed ASR/STT outputs and used only for the
descriptive phonetic-vs-omitted breakdown; they were not exhaustively
verified and may undercount Korean-phonetic renderings. The headline
English-script rendering rate depends on canonical English forms and
accepted English variants, not on the completeness of the
\texttt{korean\_phonetic} field. The \textbf{chunk-level script-consistency} measure reports, for the chunked gpt-4o conditions, the per-3-minute-chunk proportion of rendered curated domain-term tokens that appear in English/Latin rather than Korean phonetic script — that is, English / (English + phonetic) computed within the output, not a reference-normalized recall like the headline rate; chunks in which the output renders no curated domain term in either script carry no rate.

\subsection{5.2 Human annotation (single annotator, single
pass)}\label{human-annotation-single-annotator-single-pass}

Human annotation was performed by a single author-annotator in one pass,
without independent replication or adjudication.

\begin{itemize}
\tightlist
\item
  \textbf{Semantic faithfulness:} for the two post-processed conditions,
  every reference sentence was labeled with the taxonomy \{Faithful,
  Minor rewrite, Omission, Addition, Substitution, Polarity error,
  Relation error, Unclear\}. A drifted sentence received a primary drift
  label and, where applicable, one or more additional (compound) labels,
  so labels are not mutually exclusive. We distinguish \textbf{unique
  drifted rows} (distinct drifted reference sentences; the
  \texttt{drift\_rows} count) from \textbf{taxonomy label incidences}
  (per-category counts, which can exceed the number of drifted rows).
\item
  \textbf{Critical polarity:} every reference sentence carrying a
  polarity or direction cue is labeled per condition as correct / wrong
  / omitted.
\end{itemize}

All quantitative results come from author-labeled worksheets;
machine-generated draft labels were used only as non-authoritative
hints. Accordingly, rates are descriptive summaries rather than
population estimates.

\section{6. Results}\label{results}

Interpretation is descriptive throughout; with four lectures we make no
broad claims. Table 2 gives the cross-condition summary.

\begin{table}[!t]
\centering
\begin{minipage}{\linewidth}
\textbf{Table 2. Cross-condition summary.} CER, WER, and English-script
rendering are lecture-level \textbf{macro means} (each lecture weighted
equally); Korean phonetic count, polarity failure, and semantic drift
are \textbf{pooled counts} over four lectures. Semantic drift was not
annotated for the three raw-ASR conditions and is shown as ``---''.
\end{minipage}
\par\vspace{10pt}
\small
\setlength{\tabcolsep}{3pt}
\begin{tabular}{@{}
  >{\raggedright\arraybackslash}p{(\linewidth - 12\tabcolsep) * \real{0.1429}}
  >{\raggedleft\arraybackslash}p{(\linewidth - 12\tabcolsep) * \real{0.1429}}
  >{\raggedleft\arraybackslash}p{(\linewidth - 12\tabcolsep) * \real{0.1429}}
  >{\raggedleft\arraybackslash}p{(\linewidth - 12\tabcolsep) * \real{0.1429}}
  >{\raggedleft\arraybackslash}p{(\linewidth - 12\tabcolsep) * \real{0.1429}}
  >{\raggedleft\arraybackslash}p{(\linewidth - 12\tabcolsep) * \real{0.1429}}
  >{\raggedleft\arraybackslash}p{(\linewidth - 12\tabcolsep) * \real{0.1429}}@{}}
\toprule\noalign{}
\begin{minipage}[b]{\linewidth}\raggedright
condition
\end{minipage} & \begin{minipage}[b]{\linewidth}\raggedleft
CER (macro)
\end{minipage} & \begin{minipage}[b]{\linewidth}\raggedleft
WER (macro)
\end{minipage} & \begin{minipage}[b]{\linewidth}\raggedleft
English-script rendering (macro)
\end{minipage} & \begin{minipage}[b]{\linewidth}\raggedleft
Korean phonetic count (pooled)
\end{minipage} & \begin{minipage}[b]{\linewidth}\raggedleft
polarity failure (pooled)
\end{minipage} & \begin{minipage}[b]{\linewidth}\raggedleft
semantic drift (pooled)
\end{minipage} \\
\midrule\noalign{}
raw whisper-1 & 0.3128 & 0.2966 & 0.3937 & 398 & 7 & --- \\
AI\_LectureNote & 0.3879 & 0.5843 & 0.7077 & 15 & 11 & 34 \\
raw gpt-4o-transcribe & 0.3982 & 0.3067 & 0.2561 & 473 & 11 & --- \\
prompted gpt-4o-transcribe & 0.4342 & 0.3341 & 0.2198 & 429 & 9 & --- \\
gpt4o\_ailn\_post & 0.4198 & 0.6078 & 0.6468 & 17 & 13 & 36 \\
\bottomrule\noalign{}
\end{tabular}
\end{table}

\subsection{6.1 Surface metrics do not capture the
trade-off}\label{surface-metrics-do-not-capture-the-trade-off}
Macro CER did not cleanly separate the five conditions, and the lowest CER was raw whisper-1 rather than a post-processed condition. Post-processing improved surface distance mainly in lectures with heavy transliteration, but worsened it when raw ASR was already close to the study-transcript reference. This lecture-dependent pattern motivates reporting script rendering and semantic faithfulness separately rather than relying on CER/WER alone. Because the post-processing stage rewrites the ASR output rather than retranscribing the audio, the sharp increase in WER after post-processing—0.30 to 0.58 for whisper-1 and 0.31 to 0.61 for gpt-4o—offers a descriptive indication of substantial lexical rewriting introduced by post-processing, consistent with the observed surface-level drift.

\subsection{6.2 Post-processing raised English-script rendering in the
pilot}\label{post-processing-raised-english-script-rendering-in-the-pilot}

Table 2 shows that post-processing improved macro English-script
rendering on both whisper-1 (0.39→0.71) and gpt-4o (0.26→0.65) paths,
while reducing Korean phonetic occurrences from 398→15 and 473→17,
respectively. The gain was largest where raw ASR transliterated most
heavily (\texttt{diuretics\_01} rendering rate 0.05 → 0.77;
\texttt{acuteinflammation\_02} rendering rate 0.16 → 0.78), but was not
guaranteed when raw output already had a high English-script rendering
rate (\texttt{anthrax\_01} 0.56 → 0.54; \texttt{anticancerdrugs\_02}
0.81 → 0.75). Because the rendering rate measures script rather than
medical correctness, these gains must be interpreted alongside the drift
and polarity analyses in Section 6.3; a transcript can render many terms
in English while still substituting medically wrong terms.

\begin{figure}
\centering
\pandocbounded{\includegraphics[keepaspectratio,alt={Readability gain versus observed faithfulness risk, N=4. Left: macro English-script rendering by condition. Right: per-lecture semantic-drift rate for the two post-processed conditions.}]{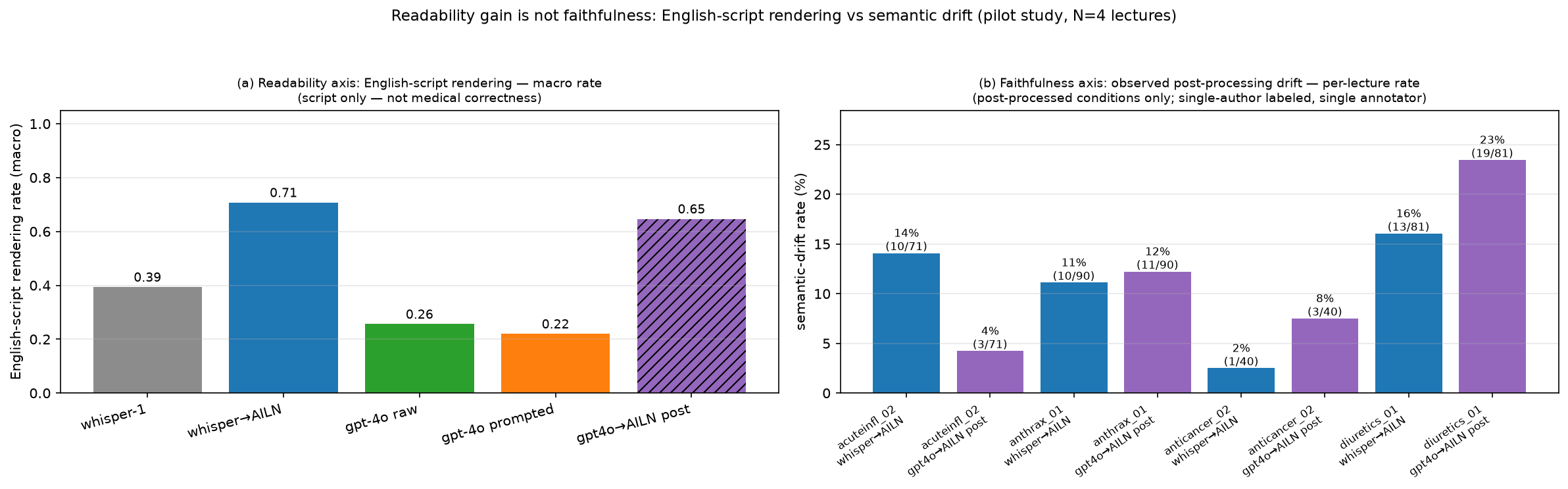}}
\caption{Readability gain versus observed faithfulness risk, N=4. Left:
macro English-script rendering by condition. Right: per-lecture
semantic-drift rate for the two post-processed conditions.}
\end{figure}

\textbf{Chunking and prompt sensitivity}. Under the 3-minute gpt-4o
chunking policy, per-chunk rendering varied sharply within lectures
(\texttt{acuteinflammation\_02} 0.94 then 0.07; \texttt{diuretics\_01}
0.00--0.29; \texttt{anticancerdrugs\_02} 0.38 then 0.96), consistent
with script-selection instability. A single minimal Korean prompt
changed rendering inconsistently across lectures (\texttt{diuretics\_01}
+0.12, \texttt{anthrax\_01} +0.13, \texttt{acuteinflammation\_02} −0.17,
\texttt{anticancerdrugs\_02} −0.22), so we treat it only as preliminary
prompt-sensitivity evidence.

\subsection{6.3 Observed faithfulness errors: semantic drift and
polarity}\label{observed-faithfulness-errors-semantic-drift-and-polarity}

\textbf{Semantic drift.} Semantic drift was annotated only for the two
post-processed conditions, so Table 3 should not be read as an aggregate
raw-vs-post comparison. Table 3 reports unique drifted reference
sentences, whereas Figure 3 reports drift-label incidences, which can
exceed unique-row counts because compound labels are allowed.

\Needspace{20\baselineskip}
\textbf{Table 3. Semantic-drift summary for the two post-processed
conditions.} Faithful/minor/drift counts are row-level annotations;
drift counts denote unique drifted reference sentences.

\begin{center}
\scriptsize
\setlength{\tabcolsep}{3pt}
\begin{tabular}{@{}
  >{\raggedright\arraybackslash}p{(\linewidth - 10\tabcolsep) * \real{0.2250}}
  >{\raggedleft\arraybackslash}p{(\linewidth - 10\tabcolsep) * \real{0.0700}}
  >{\raggedright\arraybackslash}p{(\linewidth - 10\tabcolsep) * \real{0.2050}}
  >{\raggedleft\arraybackslash}p{(\linewidth - 10\tabcolsep) * \real{0.0900}}
  >{\raggedright\arraybackslash}p{(\linewidth - 10\tabcolsep) * \real{0.2550}}
  >{\raggedleft\arraybackslash}p{(\linewidth - 10\tabcolsep) * \real{0.1550}}@{}}
\toprule\noalign{}
\begin{minipage}[b]{\linewidth}\raggedright
lecture
\end{minipage} & \begin{minipage}[b]{\linewidth}\raggedleft
labeled
\end{minipage} & \begin{minipage}[b]{\linewidth}\raggedright
AI\_LectureNote faithful/minor/\textbf{drift}
\end{minipage} & \begin{minipage}[b]{\linewidth}\raggedleft
AI drift rate
\end{minipage} & \begin{minipage}[b]{\linewidth}\raggedright
\texttt{gpt4o\_ailn\_post} faithful/minor/\textbf{drift}
\end{minipage} & \begin{minipage}[b]{\linewidth}\raggedleft
\texttt{gpt4o\_ailn\_post} drift rate
\end{minipage} \\
\midrule\noalign{}
\texttt{diuretics\_01} & 81 & 66 / 2 / \textbf{13} & 16.0\% & 56 / 6 /
\textbf{19} & 23.5\% \\
\texttt{acuteinflammation\_02} & 71 & 57 / 4 / \textbf{10} & 14.1\% & 64
/ 4 / \textbf{3} & 4.2\% \\
\texttt{anthrax\_01} & 90 & 78 / 2 / \textbf{10} & 11.1\% & 76 / 3 /
\textbf{11} & 12.2\% \\
\texttt{anticancerdrugs\_02} & 40 & 38 / 1 / \textbf{1} & 2.5\% & 37 / 0
/ \textbf{3} & 7.5\% \\
\textbf{pooled total} & 282 & \textbf{34} & --- & \textbf{36} & --- \\
\bottomrule\noalign{}
\end{tabular}
\end{center}

Semantic drift was observed in both post-processed conditions, with
similar pooled counts but different lecture-level distributions. Table 3
therefore supports treating drift as workflow- and input-sensitive
rather than as a single aggregate rate. The redistribution of affected
rows motivates the cross-input analysis in Section 6.4.

\begin{figure}
\centering
\pandocbounded{\includegraphics[keepaspectratio,alt={Semantic-drift taxonomy for the two post-processed conditions. Bars show drift-label incidences by lecture and error type; labels are not mutually exclusive, so category counts can exceed unique drifted-row totals.}]{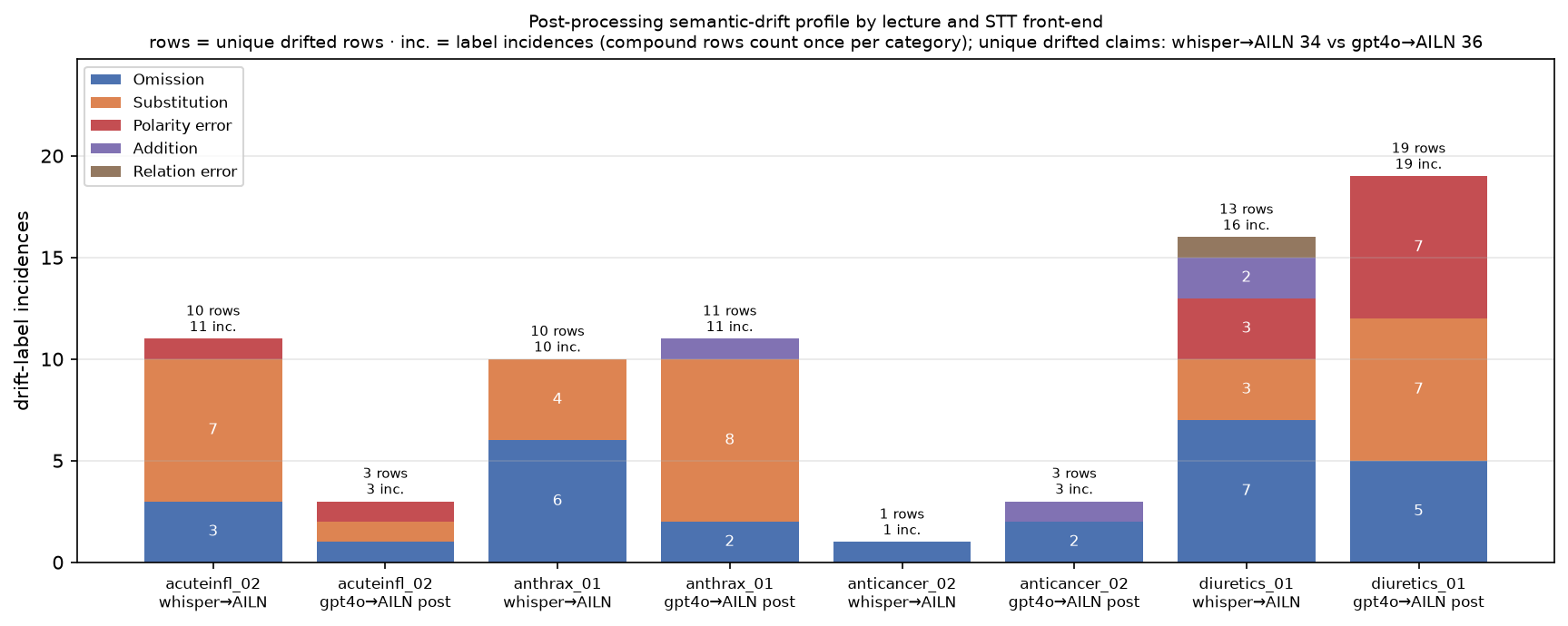}}
\caption{Semantic-drift taxonomy for the two post-processed conditions.
Bars show drift-label incidences by lecture and error type; labels are
not mutually exclusive, so category counts can exceed unique drifted-row
totals.}
\end{figure}

\textbf{Critical polarity.} We report \texttt{wrong} and
\texttt{omitted} separately but interpret a polarity failure as the
\texttt{wrong\ +\ omitted} set, pooled over 101 polarity-cue rows.

\Needspace{12\baselineskip}
\textbf{Table 4. Critical-polarity failures by condition.} Counts are
pooled over 101 polarity-cue rows; failure is defined as wrong plus
omitted.

\begin{center}
\small
\begin{tabular}{@{}lrrr@{}}
\toprule\noalign{}
condition & wrong & omitted & \textbf{failure (= w + o)} \\
\midrule\noalign{}
raw whisper-1 & 7 & 0 & \textbf{7} \\
AI\_LectureNote & 8 & 3 & \textbf{11} \\
raw gpt-4o-transcribe & 11 & 0 & \textbf{11} \\
prompted gpt-4o & 9 & 0 & \textbf{9} \\
gpt4o\_ailn\_post & 9 & 4 & \textbf{13} \\
\bottomrule\noalign{}
\end{tabular}
\end{center}

\begin{figure}
\centering
\pandocbounded{\includegraphics[keepaspectratio,alt={Critical medical polarity, N=4. Panel (a) shows wrong and omitted polarity-cue rows by condition; panel (b) shows the whisper-1 → AI\_LectureNote polarity transition for diuretics\_01.}]{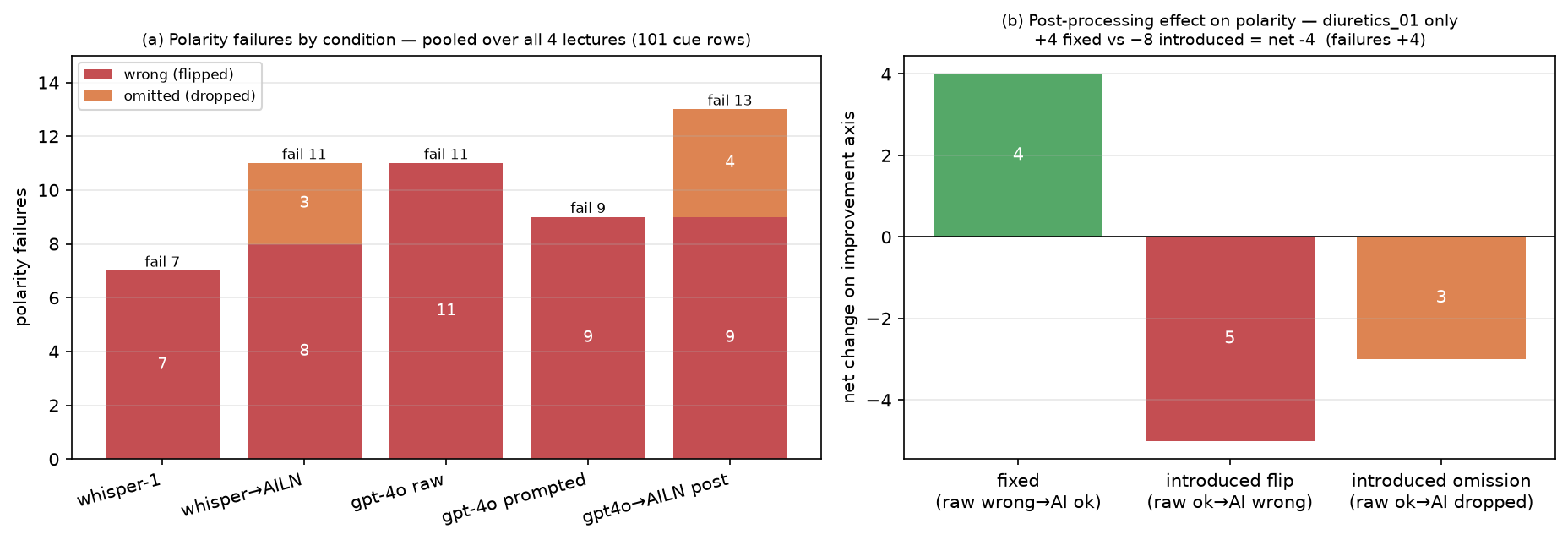}}
\caption{Critical medical polarity, N=4. Panel (a) shows wrong and
omitted polarity-cue rows by condition; panel (b) shows the whisper-1 →
AI\_LectureNote polarity transition for \protect\texttt{diuretics\_01}.}
\end{figure}

Polarity failures were rarer than general semantic drift but more
clinically salient, and they concentrated in \texttt{diuretics\_01}.
Counting omissions together with wrong polarity changed the
interpretation of post-processing performance: AI\_LectureNote
introduced omissions that would be missed by a wrong-only count.

\subsection{6.4 Cross-input control: introduced versus propagated
errors}\label{cross-input-control-introduced-versus-propagated-errors}

To examine why similar pooled drift counts produced different affected
sentences, the cross-input control probes two candidate sources of
post-processed errors. We label selected errors as \textbf{introduced}
when raw ASR preserved the relevant information but post-processing lost
or changed it, and \textbf{propagated} when post-processing carried
forward an upstream ASR error. Representative introduced errors included
polarity or term substitutions such as ``hypercalcemia'' →
``hypercalciuria,'' whereas propagated errors included upstream
substitutions such as ``lymphocyte'' → ``lipocyte.'' Additional examples
are listed in Appendix B.

The two post-processed conditions showed different cross-input patterns
in this pilot. For directional polarity, the two conditions agreed on 95
of 101 rows, with a failure-set Jaccard of 0.60 (9 shared of 15 unioned
failures). Several rows that all
raw front-ends transcribed correctly flipped to the same wrong value
after post-processing, consistent with a recurring
post-processing-associated polarity pattern. For general semantic drift,
the overlap was much lower: the drift-set Jaccard was 0.23, with 13
shared drifted claims out of a union of 57. Aggregate drift was similar
across the two post-processed conditions (34 vs.~36), but the affected
sentences shifted with the input ASR.

This input sensitivity was clearest in \texttt{acuteinflammation\_02}:
applying the same post-processing to gpt-4o output reduced drift from 10
to 3 and substitutions from 7 to 1, consistent with fewer propagated
whisper-path substitutions. At the same time, shared drift remained
across front-ends, especially in the \texttt{diuretics\_01}
mechanism/electrolyte cluster and \texttt{anthrax\_01} compression
omissions. Thus, the control argues against treating the observed gains
and errors as merely whisper-1 artifacts: applying the same
post-processing to raw gpt-4o also raised English-script rendering (0.26
→ 0.65), but did not eliminate observed faithfulness errors. These
patterns should be interpreted as workflow-specific.

\section{7. Discussion}\label{discussion}

Four points follow from the pilot.

First, student-facing transcript utility is not the same as generic STT
accuracy. The five conditions trade off differently across surface
accuracy, term-script rendering, chunk consistency, and medical-meaning
preservation; CER/WER alone miss this trade-off. The original
transliteration problem persisted in some lectures, including under the
chunked gpt-4o condition, but not uniformly. The clearest counterexample
is \texttt{anticancerdrugs\_02}, where raw whisper-1 already rendered
0.81 of curated terms in English script; this is also the shortest lecture
(4m30s), so its high raw rendering may reflect length as well as content.

Second, English-script rendering is a useful readability proxy for
Korean--English medical lectures, but it is not semantic faithfulness. A
high rendering rate can coexist with wrong English terms (Section 6.3),
so script rendering and medical meaning must be measured separately.

Third, post-ASR rewriting can improve readability while adding
paraphrase, compression, and faithfulness risks, especially when raw ASR
is already close to the reference. The cross-input comparison suggests
that some polarity failures were post-processing-associated, whereas
broader semantic drift was more input-sensitive.

Fourth,  these findings should not be read as evidence that all domain-aware
post-processing introduces semantic drift. AI\_LectureNote is a
readability-oriented LLM rewriting pipeline rather than a constrained
term-normalization system; the diagnostic point is that script-rendering
gains and semantic faithfulness should be evaluated separately.

\section{8. Limitations}\label{limitations}

This retrospective pilot bounds every claim above; we make no benchmark,
population-rate, educational, or clinical claim.

\begin{itemize}
\tightlist
\item
  \textbf{Scale and confounding.} Four lectures only. Speaker, topic,
  lecture length, and reference style remain confounded, even after
  adding a second cdh lecture.
\item
  \textbf{Annotation provenance and evaluator overlap.} References and
  labels were created via single-pass annotation by the system
  developers (single author-annotator) with no second pass, independent
  replication, adjudication, or inter-annotator agreement. Furthermore,
  references are lightly cleaned study transcripts rather than
  gold-standard verbatim transcripts.
\item
  \textbf{Polarity concentration.} Polarity failures concentrate in
  \texttt{diuretics\_01}, so polarity comparisons are illustrative
  rather than robust rate estimates.
\item
  \textbf{Implementation choices.} Condition 5 is a diagnostic control,
  not an optimized modern-STT post-processor; the gpt-4o conditions use
  3-minute non-overlapping chunks and one minimal prompt, and the
  whisper-1 condition is a 2026 rerun rather than a 2023 snapshot.
\item
  \textbf{Metric scope.} English-script rendering is recall-like and
  script-only; it does not measure term correctness or hallucinated
  terms except through missed reference terms.
\item
  \textbf{Korean-phonetic diagnostic counts.} The Korean-phonetic counts
  are diagnostic lower-bound counts based on a machine-assisted list of
  observed phonetic variants; the phonetic variant lists were not
  exhaustively verified.
\end{itemize}

\section{9. Conclusion}\label{conclusion}

Across four author-recorded lectures, AI\_LectureNote-style
post-processing improved English-script rendering on both input paths,
but these readability gains coexisted with semantic drift and polarity
failures. The cross-input control suggested that polarity failures were
more stable across raw front-ends than general semantic drift, whose
affected sentences shifted with the input ASR. The central
methodological takeaway is that surface distance, term-script rendering,
chunk-level consistency, and medical-meaning preservation should be
evaluated separately. Future work should test more constrained
alternatives, such as span-level glossary replacement, verifier-backed
correction, or constrained rewriting.

\Needspace{20\baselineskip}
\section{10. Reproducibility / Data and Code
Availability}\label{reproducibility-data-and-code-availability}

From the \texttt{evaluation\_paper/} directory, all tables and figures
regenerate with \texttt{python\ analysis/run\_all.py}. Source files are
organized as (paths relative to \texttt{evaluation\_paper/}):

\begin{itemize}
\tightlist
\item
  \textbf{References}\\
  \mbox{\texttt{data/\textless{}lecture\textgreater{}/reference.txt}}
\item
  \textbf{Semantic labels}\\
  \mbox{\texttt{annotations/\textless{}lecture\textgreater{}/semantic\_review.csv}}
\item
  \textbf{Condition-5 semantic labels}\\
  \mbox{\texttt{annotations/\textless{}lecture\textgreater{}/semantic\_review\_gpt4o\_ailn\_post.csv}}
\item
  \textbf{Polarity labels}\\
  \mbox{\texttt{annotations/\textless{}lecture\textgreater{}/polarity\_review.csv}}
\item
  \textbf{Domain terms}\\
  \mbox{\texttt{annotations/\textless{}lecture\textgreater{}/domain\_terms.csv}}
\item
  \textbf{Condition manifest}\\
  \mbox{\texttt{manifest/condition\_manifest.csv}}
\end{itemize}

The condition manifest records model names, API dates, prompt text, and
the condition-5 pipeline description.

Code, speaker-coded evaluation transcripts, references, annotation
files, and generated tables/figures are released in the public
repository:
\url{https://github.com/boyskier/ailecturenote-retrospective-pilot}.
Where audio is included, it is released with explicit author-speaker
consent for research reproducibility; because the audio contains
identifiable voices, it should not be treated as anonymized.

\section{Acknowledgements}\label{acknowledgements}

We thank Professor Hogun Park for supervising the 2024 CO-Deep Learning
student project that laid the foundation for this work.

\section{Appendix A. Supplementary
figures}\label{appendix-a.-supplementary-figures}

The four main figures appear inline (Figure 1, Section 4.1; Figure 2,
Section 6.2; Figure 3 and Figure 4, Section 6.3). The following
supplementary figures provide per-lecture detail and are reproduced by
\texttt{python\ analysis/run\_all.py}:

\begin{itemize}
\tightlist
\item
  \textbf{Figure S1} (\path{figures/fig2_cer_wer.png}) --- per-lecture
  CER (primary) and WER (secondary) for all five conditions, showing the
  \texttt{anthrax\_01} and \texttt{anticancerdrugs\_02} divergence where
  raw whisper-1 is clean but post-processing raises CER (Section 6.1).
\item
  \textbf{Figure S2} (\path{figures/fig1_english_script_rate.png}) ---
  English-script rendering rate by condition and lecture, the
  per-lecture detail behind the headline trade-off in Figure 2 (Section
  6.2).
\item
  \textbf{Figures S3a--d} (\texttt{figures/fig3\_chunk\_\textless{}lecture\textgreater{}.png}) ---
  per-3-minute-chunk English vs.~Korean-phonetic counts for raw and
  prompted gpt-4o, one panel per lecture
  (\texttt{acuteinflammation\_02}, \texttt{anthrax\_01},
  \texttt{anticancerdrugs\_02}, \texttt{diuretics\_01}), illustrating
  the chunk-level script instability (Section 6.2).
\end{itemize}

\section{Appendix B. Illustrative
errors}\label{appendix-b.-illustrative-errors}

Appendix B lists illustrative author-reviewed errors. The examples are drawn from
\texttt{table7\_semantic\_drift\_examples.csv}. Examples include propagated substitutions such as lymphocyte $\rightarrow$ ``lipocyte''
and tissue edema $\rightarrow$ ``tissue adenoma''; introduced or
post-processing-associated polarity/substitution errors such as hypercalcemia
$\rightarrow$ ``hypercalciuria'' and hypokalemia-prevention $\rightarrow$
``hyperkalemia-prevention''; and additions such as an unmentioned ``ibuprofen''
in the ototoxicity discussion. Condition 5 also produced a
concept error in \texttt{anticancerdrugs\_02}, conflating antimetabolites
with alkylating agents.

For balance, post-processing also corrected some upstream errors in
\texttt{anthrax\_01}, including restoring whisper-1's ``Annihilation''
to ``Inhalation'', preserving the 90--95\% / 20\% mortality figures, and
restoring the PA/EF/LF abbreviations in English. Thus, the examples
illustrate mixed effects of readability-oriented rewriting rather than a
one-directional degradation.

\begin{center}\rule{0.5\linewidth}{0.5pt}\end{center}

\section{References}\label{references}

\begin{itemize}
\tightlist
\item
  Adedeji, A., Joshi, S., \& Doohan, B. (2024). \emph{The Sound of
  Healthcare: Improving Medical Transcription ASR Accuracy with Large
  Language Models.} arXiv:2402.07658.
\item
  Adedeji, A., Sanni, M., Ayodele, E., Joshi, S., \& Olatunji, T.
  (2025). \emph{The Multicultural Medical Assistant: Can LLMs Improve
  Medical ASR Errors Across Borders?} arXiv:2501.15310.
\item
  Chapman, W. W., Bridewell, W., Hanbury, P., Cooper, G. F., \&
  Buchanan, B. G. (2001). \emph{A simple algorithm for identifying
  negated findings and diseases in discharge summaries.} Journal of
  Biomedical Informatics, 34(5), 301--310. doi:10.1006/jbin.2001.1029.
\item
  Ma, R., Qian, M., Manakul, P., Gales, M., \& Knill, K. (2023).
  \emph{Can Generative Large Language Models Perform ASR Error
  Correction?} arXiv:2307.04172.
\item
  OpenAI. (2025). \emph{Speech to text.} OpenAI API documentation.
  Accessed 2026-06-27.
  https://developers.openai.com/api/docs/guides/speech-to-text
\item
  Paik, G., Kim, Y., Lee, S., Ahn, S., \& Kim, C. (2025). \emph{HiKE:
  Hierarchical Evaluation Framework for Korean--English Code-Switching
  Speech Recognition.} arXiv:2509.24613.
\item
  Peng, Y., Wang, X., Lu, L., Bagheri, M., Summers, R. M., \& Lu, Z.
  (2017). \emph{NegBio: a high-performance tool for negation and
  uncertainty detection in radiology reports.} arXiv:1712.05898.
\item
  Radford, A., Kim, J. W., Xu, T., Brockman, G., McLeavey, C., \&
  Sutskever, I. (2023). \emph{Robust speech recognition via large-scale
  weak supervision} (Whisper). Proceedings of the 40th International
  Conference on Machine Learning, PMLR 202:28492--28518.
  arXiv:2212.04356.
\item
  Sitaram, S., Chandu, K. R., Rallabandi, S. K., \& Black, A. W. (2019).
  \emph{A survey of code-switched speech and language processing.}
  arXiv:1904.00784.
\item
  Sudo, Y., Shakeel, M., Fukumoto, Y., Peng, Y., \& Watanabe, S.
  (2024a). \emph{Contextualized Automatic Speech Recognition with
  Attention-Based Bias Phrase Boosted Beam Search.} arXiv:2401.10449.
\item
  Sudo, Y., Fukumoto, Y., Shakeel, M., Peng, Y., \& Watanabe, S.
  (2024b). \emph{Contextualized Automatic Speech Recognition with
  Dynamic Vocabulary.} arXiv:2405.13344.
\item
  Tang, Z., Wang, D., Zhou, Z., Liu, Y., Huang, S., \& Shang, S. (2025).
  \emph{Chain of Correction for Full-text Speech Recognition with Large
  Language Models.} arXiv:2504.01519.
\item
  Uzuner, Ö., South, B. R., Shen, S., \& DuVall, S. L. (2011).
  \emph{2010 i2b2/VA challenge on concepts, assertions, and relations in
  clinical text.} Journal of the American Medical Informatics
  Association, 18(5), 552--556. doi:10.1136/amiajnl-2011-000203.
\item
  Wang, J., Kim, J., Kim, S., \& Lee, Y. (2019). \emph{Exploring
  Lexicon-Free Modeling Units for End-to-End Korean and Korean--English
  Code-Switching Speech Recognition.} arXiv:1910.11590.
\item
  Zheng, X., Dong, S., Phukon, B., Hasegawa-Johnson, M., \& Yoo, C. D.
  (2026). \emph{Towards Robust Dysarthric Speech Recognition: LLM-Agent
  Post-ASR Correction Beyond WER.} arXiv:2601.21347.
\end{itemize}

\end{document}